\documentclass{ecai}
\usepackage{times}
\usepackage{cite}
\usepackage{graphicx}
\usepackage{latexsym}
\usepackage{amsfonts}
\usepackage{verbatim}
\usepackage{amsmath}
\usepackage{amsfonts}
\usepackage{amssymb}
\usepackage{booktabs} 
\usepackage{algorithm}
\usepackage{graphicx}
\usepackage{subfigure}
\usepackage{algpseudocode}  
\newtheorem{definition}{Definition}

\begin{document}

\title{EPNE: Evolutionary Pattern Preserving \\ Network Embedding}

\author{Junshan Wang\institute{Key Laboratory of Machine Perception, Ministry of Education, Peking University, China, email: wangjunshan@pku.edu.cn. \textsuperscript{*} These authors contributed equally to the work.} \textsuperscript{ *} \and Yilun Jin\institute{The Hong Kong University of Science and Technology, Hong Kong SAR, China, email: yilun.jin@connect.ust.hk. \textsuperscript{*} These authors contributed equally to the work.} \textsuperscript{ *} \and Guojie Song\institute{Key Laboratory of Machine Perception, Ministry of Education, Peking University, China, email: gjsong@pku.edu.cn. \textsuperscript{$\dagger$} Corresponding author.} \textsuperscript{ $\dagger$} \and Xiaojun Ma\institute{Key Laboratory of Machine Perception, Ministry of Education, Peking University, China, email: mxj@pku.edu.cn.}}

\maketitle
\bibliographystyle{ecai}

\begin{abstract}
Information networks are ubiquitous and are ideal for modeling relational data. Networks being sparse and irregular, network embedding algorithms have caught the attention of many researchers, who came up with numerous embeddings algorithms in static networks. Yet in real life, networks constantly evolve over time. Hence, evolutionary patterns, namely how nodes develop itself over time, would serve as a powerful complement to static structures in embedding networks, on which relatively few works focus. In this paper, we propose EPNE, a temporal network embedding model preserving evolutionary patterns of the local structure of nodes. In particular, we analyze evolutionary patterns with and without periodicity and design strategies correspondingly to model such patterns in time-frequency domains based on causal convolutions. In addition, we propose a temporal objective function which is optimized simultaneously with proximity ones such that both temporal and structural information are preserved. With the adequate modeling of temporal information, our model is able to outperform other competitive methods in various prediction tasks.
\end{abstract}

\section{Introduction}
Information networks are present everywhere due to its vivid depiction of relational data. Network data being sparse and irregular, network embedding algorithms \cite{DBLP:journals/corr/abs-1711-08752}, mapping nodes to vectors such that network properties are maintained, alleviate such drawbacks and further facilitate numerous prediction tasks. Most existing embedding methods generally consider local and global node proximity, among which models including DeepWalk \cite{perozzi2014deepwalk} and GraphSAGE \cite{hamilton2017inductive} are highly popular and efficient. However, one unrealistic assumption all of the above models make is that the network is static and all nodes have been sufficiently represented through the static network structure.

In reality, nevertheless, this is hardly the case. Hardly any networks in real life remain stagnant forever, with its structure constantly evolving over time \cite{holme2015modern}. Hence, in addition to structures, temporal evolutionary patterns for nodes, namely how local structures of a node change over time would serve as complements to static structures and are indicative of key properties of nodes. Complex as such patterns are, many of them are characterized by strong periodicity. For example, in social networks, colleagues interact more frequently during weekdays, while families are more active during weekends, where interaction patterns shed light on relationships underlying edges. There are also non-periodic patterns which still illustrate certain trends. For example, a user who is starting to join a community is generally expected to further join the community, where his evolution is monotonous but shows a trend he is inclined to follow. Therefore, modeling evolutionary patterns for nodes and incorporating them into learning node embeddings will be helpful to infer node labels and identify edge relations.

\textit{Temporal Network Embedding}, reflecting the need for modeling such permanent change, came into life. Most existing temporal network embedding models either focus on certain network evolutionary processes, or are able to efficiently update their representation vectors so that they constantly embody up-to-date information. For example, DynamicTriad \cite{zhou2018dynamic} and HTNE \cite{zuo2018embedding}, both of which fall into the former category, model the triadic closure process and neighborhood formation process, respectively. However, both of them only focus on a relatively specific dynamic process of networks, and thus being unable to generalize to more complex evolutionary patterns. 

While the latter category has received wide attention \cite{li2017attributed,zhu2016scalable,du2018dynamic}, there have been relatively few efforts into modeling network evolution. Therefore, to complement the scarcity of such works, we focus on the former category of temporal network embedding, hoping to capture intricate, both periodic and non-periodic patterns. Consequently, two major challenges arise:
\begin{itemize}
    \item Temporal features, which capture evolutionary patterns of nodes, are hard to learn. On one hand, node interactions are often noisy and sparse, making it difficult for us to extract information from. On the other hand, evolutionary patterns cover a wide range of contents, including neighborhood formation and proximity drifts, consisting of complex periodic patterns with multiple frequencies as well as diverse non-periodic ones. 

    \item Straightforward methods of incorporating temporal features to representations, such as concatenation and addition, compromise the quality of both. On one hand, they come from different spaces and cannot be easily aligned. On the other hand, such operations treat temporal and structural features as separate without utilizing one to complement the other. 
\end{itemize}

To address these challenges, in this paper, we propose a temporal network embedding model, abbreviated EPNE, which combines both structural information and temporal features. The model consists of two components. First, we analyze that evolutionary patterns of nodes' local structures consist of periodic and non-periodic patterns, for which we design methods to capture in time-frequency domains based on causal convolutions. What is more, we designed an objective function that is not only able to capture node proximity, but is also able to preserve the learned evolutionary patterns through representation vectors, enabling us to jointly optimize the model to capture both properties of networks. 

To summarize, we make the following contributions:
\begin{itemize}
    \item We propose EPNE, a network embedding algorithm on temporal networks, which preserves both evolutionary patterns and topological structures of nodes.
    \item We analyze the evolutionary patterns of the local structure of nodes, based upon which, a novel strategy is designed to learn temporal features for periodic and non-periodic patterns using causal convolutions.
    \item We evaluate our model on several real-world networks for node and edge classification. The results demonstrate that our model is capable of preserving the temporal features of nodes and outperforms its counterparts.
\end{itemize}

\section{Related Work}
\textbf{Static Network Embedding}.  With the advent of Skip-gram \cite{mikolov2013distributed} models in natural language processing, similar models on graphs came into being, among which DeepWalk \cite{perozzi2014deepwalk},  Node2Vec \cite{grover2016node2vec} and LINE \cite{tang2015line} are popular models, not only because they show impressive performance, but also because of the large number of algorithms derived upon them \cite{yang2015network,long2019hierarchical}. Later, with the spread of deep learning, deep models \cite{zhou2018graph,xu2018powerful} including GraphSAGE \cite{hamilton2017inductive} and GCN \cite{kipf2016semi} were developed to perform deep learning for network embedding. 

\noindent \textbf{Temporal Network Embedding}. Existing works mainly focus on two aspects illustrating network dynamics. On one hand, online update methods, through which representation vectors can be efficiently adjusted to reflect the latest changes in network topology, have been widely studied \cite{li2017attributed,zhu2016scalable,du2018dynamic,ma2018depthlgp}, demonstrating comparable results and alleviating the need to retrain the model. On the other hand, network evolving processes and factors leading to them have also received extensive attention, typical examples of which are DynamicTriad, HTNE, CTDNE \cite{nguyen2018continuous}, tNodeEmbed \cite{singer2019node}, DynamicGCN \cite{pareja2019evolvegcn} and most recently, HierTCN \cite{you2019hierarchical}, and have all achieved outstanding performance in various tasks like link prediction, and visualizing dynamics. 

DynamicTriad \cite{zhou2018dynamic} models the triadic closure process ubiquitous in social networks, where two nodes sharing a neighbor in common are motivated to form links. 
HTNE \cite{zuo2018embedding} models the neighborhood formation sequences using Hawkes Process, through which the whole local structure of nodes is incorporated with temporally-aware weights.
tNodeEmbed \cite{singer2019node} present a joint loss that creates a temporal embedding of a node using LSTM to combine its historical temporal embeddings.
EvolveGCN \cite{pareja2019evolvegcn} adapts a GCN model along the temporal dimension and captures the dynamics of the graph sequence by using an RNN to evolve the GCN parameters.
Yet elaborate as all of them are, only specific dynamic processes are taken into account, to which our model is trying to complement. Specifically, all of them fail to account for periodic temporal patterns, which we will show to be highly indicative.

\begin{figure*}[!htbp]
\centering
\includegraphics[width=0.9\linewidth]{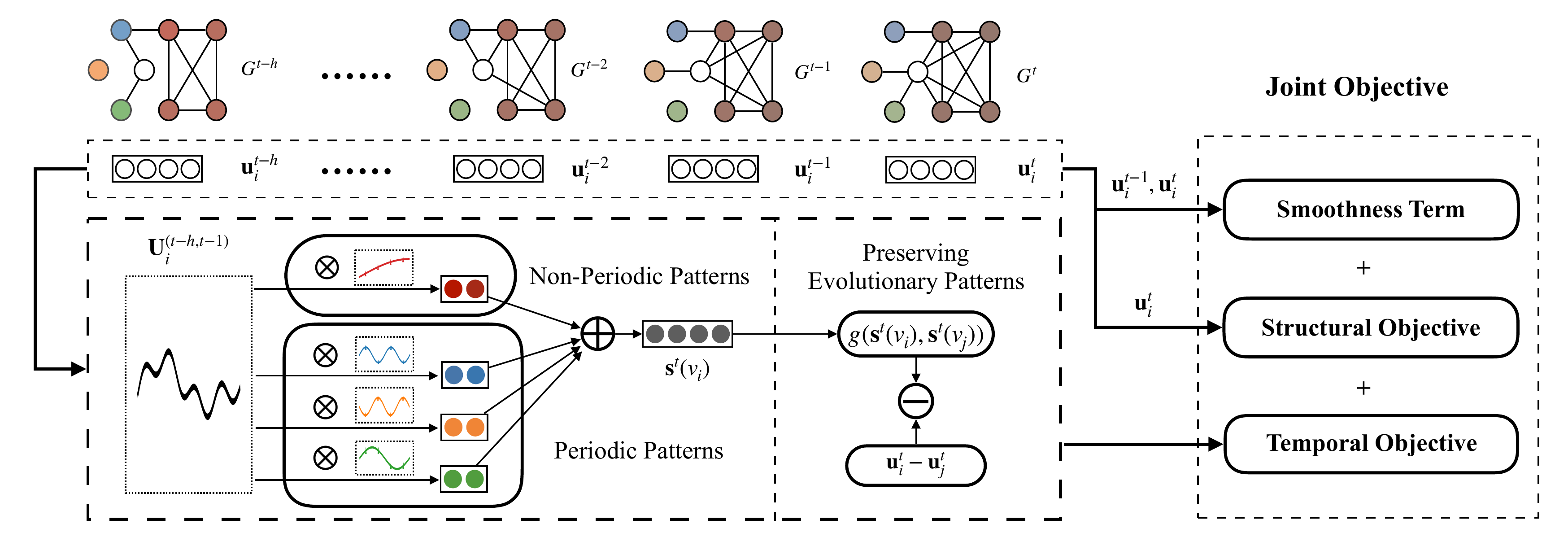}
\caption{Illustration of our Evolutionary Pattern preserving Network Embedding, abbreviated EPNE that learns embeddings $\mathbf{u}_i^t$ of node $v_i$ at time $t$. 
$\otimes$, $\oplus$ and $\ominus$ represent convolution, concatenation and operations measuring distance respectively.
The block on the middle depicts our model preserving multi-scaled, periodic and non-periodic evolutionary patterns. The block on the right describes the joint objective of our model consisting of smoothness term, structural and temporal objective.}
\label{fig:model}
\end{figure*}

\section{Preliminaries}
In this section, we define our problems in the context of temporal networks, followed by the introduction of causal convolutions, which will be used to learn temporal features of nodes in the next section.  

\subsection{Problem Definition}
\begin{definition}
A \textbf{Temporal Network} is defined as
\begin{equation}
 G = \{G^1,...,G^T\}, 
\end{equation}
where $G^t = (V, E^t)$ represents the network snapshot at time $t$. $V=\{v_i\}, i=1,2,...,|V|$ represents the set of nodes and $E^t=\{e_{ij}^t|i,j\in V\}$ represents the set of edges at time $t$. $E=\bigcup_{t \le T} E^t$ denotes the set of \textbf{static edges}, i.e. the set of edges that exist at least once in all time steps. $e_{ij}^t$ indicates an (undirected) edge between $v_i$ and $v_j$ at time $t$. We denote $N^t(v_i)$ as the context of $v_i$ in a random walk at time $t$.
\label{defn:tempnet}
\end{definition}

\begin{definition}
Given a temporal network $G = \{G^1,...,G^T\}$, the problem of \textbf{Temporal Network Embedding} aims to learn a mapping function
\begin{equation}
f^t:v_i \to \mathbf{u}^t_i\in \mathbb{R}^{d},
\end{equation}
where $d \ll |V|$ and $\mathbf{u}_i^t$ preserves both local network structure and evolutionary patterns of node $v_i$ at time $t$.
\end{definition}

While DynamicTriad and HTNE capture temporal evolutions of nodes by modeling its interaction with neighbors, we consider it inadequate for two reasons. On one hand, higher-order proximity is generally modeled by network embeddings, while interactions with neighbors fail to take such higher-order information into account. On the other hand, interactions between nodes can be noisy and sparse, which compromises the quality of our learned features. 

On the contrary, the representation vector of a node is able to higher-order structural information and consequently, its changes reflect a richer notion of temporal drifts. What is more, they do not suffer from sparsity and are generally less noisy. Therefore, we focus on the sequence of representation vectors in a fixed history length $h$
\begin{equation}
\textbf{U}_i^{(t-h,t)} = (\textbf{u}_i^{t-h}, \textbf{u}_i^{t-h+1}, ..., \textbf{u}_i^{t-1}),
\end{equation}
to learn the temporal features of a node.

\subsection{Causal Convolutions}
Recurrent Neural Networks (RNNs) are popular approaches for modeling sequential data, but they are typically slow to train due to recurrent connections. Hence we resort to causal convolutions \cite{DBLP:journals/corr/abs-1803-01271}, a comparably flexible but much more efficient architecture for modeling sequential data, whose computation also follows the sequence as provided by the data. In temporal networks, the causal convolution for sequence $\mathbf{U}_i^{(t-h,t)}$ is
\begin{equation}
\mathbf{s}^t(v_i) = \mathbf{U}_i^{(t-h,t)} \cdot \mathbf{f} = \sum_{k=0}^{h-1} \mathbf{u}_i^{t-h+k} \mathbf{f}(k),
\end{equation}
where $\mathbf{f} \in \mathbb{R}^h$ is the convolution kernel. Conventionally, the convolution kernel $\mathbf{f}$ are trainable parameters, but aimed at our problem, we design fixed convolution kernels $\mathbf{f}$ with prior analysis, so that our model can learn temporal features of nodes in temporal networks more efficiently. 

\section{Proposed Model}
In this section, we describe our model for temporal network embedding, which preserves both evolutionary patterns and network topology in representation vectors, as illustrated in Figure  \ref{fig:model}. First, as evolutionary patterns of local structures consist of periodic patterns and non-periodic ones, we design a method to learn temporal features capturing evolutionary patterns in time-frequency domains. Second, in order to map both evolutionary patterns and network structure into the vector space, we design a temporal objective function and optimize it along with structural objective jointly.

\subsection{Modeling Evolutionary Patterns}
In this section, we introduce how temporal features representing evolutionary patterns are made up of, as well as how they are extracted and processed in our model. Primarily, we consider evolutionary patterns from two aspects.
\begin{itemize}
    \item \textbf{Periodic evolutionary patterns.} Local structures may emerge and disappear with strong regularity. For example, sports events like the Super Bowl is held annually, and related Reddit submissions will accordingly follow such regularity. 
    \item \textbf{Non-periodic evolutionary patterns.} Local structures and interactions may also follow non-periodic but gradual evolutionary patterns. For example, a user who has just watched a concert of Beethoven music will be inclined to view that of other musicians, such as Mozart and Haydn.
\end{itemize}
Aiming at modeling both patterns, we learn temporal features in frequency and time domains, respectively, before combining them to get the complete temporal features.

\subsubsection{Frequency Domain Features}
Considering node $v_i$ and one of its neighbors $v_j$ at time $t_1$ and $t_2=t_1+T$, if $\mathbf{u}_i^{t_1} \cdot \mathbf{u}_j^{t_1} \approx \mathbf{u}_i^{t_2} \cdot \mathbf{u}_j^{t_2}$, then a periodicity of $T$ between $v_i$ and $v_j$ is indicated. Periodic evolutionary patterns are critical in identifying relationships between nodes. For example, in social networks, the frequency of interactions between colleagues peak during weekdays and decline during weekends, while that between families and friends is completely the opposite, shrinking in weekdays and peaking during holidays. In co-author networks, experts from different domains may exhibit different patterns of cooperation due to different frequencies at which conferences are held. 

Inspired by ideas in signal analysis, we extract periodic temporal features by decomposing them into basic patterns $x(t) = \sum_{m=1}^M \omega_m x_m(t)$, where  $x_m(t)$ is a basic pattern and $\omega_m$ is the corresponding intensity which will serve as features in our model. 

Based on the general ideas above, we define a generating function $\mathbf{f}^*$ to generate a set of functions $\{\mathbf{f}_{a, b}\}$ as basic patterns with different scales through translating and retracting
\begin{equation}
\mathbf{f}_{a,b}(t) = \mathbf{f}^*\left(\frac{t-a}{b}\right),
\end{equation}
where $a, b$ are the shifting and scaling factors respectively. We set the number of scales as $L$. These basic patterns are regarded as convolution kernels, which are used to carry out causal convolutions to extract decomposed features as
\begin{equation}
\mathbf{s}_{f,a,b}^t(v_i) = \mathbf{U}_i^{(a,a+b)} \cdot \mathbf{f}_{a,b} = \sum_{k=t-h}^{t-1} \mathbf{u}_i^k f_{a,b}(k).
\end{equation}
We concatenate the vectors $\mathbf{s}_{f,a,b}^t(v_i)$ on different scales and shifts to get the frequency domain features of a node 
\begin{equation}
\mathbf{s}_{freq}^t(v_i) = \oplus_{a, b} \mathbf{s}^t_{f, a, b}(v_i).
\label{equ:freq_features}
\end{equation}

For the choice of generating functions, one simple yet commonly used option is Haar wavelets, a sequence of rescaled square-shaped signals used in computer vision and time series analysis for feature extraction \cite{mulcahy1997image,chan1999efficient}. Such common solutions prove to be sufficient by the experimental results for extracting general evolutionary patterns. For the shifting and scaling factors $a$ and $b$, we adopt the setting of 
\begin{align}
    b_l &= \frac{1}{2}b_{l-1},\ a_{l, k} = a_{l, k-1} + b_l,\  l= 1, 2...L,\ k\le l,
\end{align}
 as stated in \cite{chui2016introduction} to be generally adopted for Haar wavelets. In addition to common settings, our model allows for personalized designs that meet the properties of individual datasets. For instance, in social networks, we divide each time slice as a one-hour interval, and designed 
$b_0 = 1, b_1 = \frac{1}{24}, b_2 = \frac{1}{7\cdot 24}$
such that basic patterns thus generated would capture daily and weekly dynamics. 

\subsubsection{Time Domain Features}
In addition to periodic patterns, the local structure of a node may follow a steady, monotonous trend instead of periodicity. We model non-periodic evolutionary patterns by learning features in time domains based on decay kernels. Generally for humans, more recent behaviors exert a stronger impact on the present, and so does decay kernels operate on vector sequences. We define a decay kernel $\mathbf{f}_\alpha(t) = e^{-\alpha t}$, where $\alpha$ is the decay rate, which is then fed into causal convolutions on sequences $\mathbf{U}_i^{(t-h, t)}$ to capture non-periodic trends. Specifically, we have the time domain features similarly as
\begin{equation}
\mathbf{s}^t_{time}(v_i) = \mathbf{U}_i^{(t-h,t)} \cdot \mathbf{f}_\alpha.
\label{equ:time_features}
\end{equation}

We concatenate both periodic and non-periodic features to acquire the complete temporal features 
\begin{equation}
\mathbf{s}^t(v_i) = \mathbf{s}^t_{time}(v_i) \oplus \mathbf{s}^t_{freq}(v_i).
\label{equ:temporal_features}
\end{equation}
The challenge yet to overcome is that, how should the aforementioned features be incorporated to elevate the performance of the embedding algorithm. 
\subsubsection{Discussion} Here we show a brief discussion about our model and DynamicTriad, HTNE. It is evident that both DynamicTriad and HTNE only capture non-periodic temporal features in that, closed triads will not be open again and the intensity of Hawkes processes is monotonously decreasing. By comparison, we propose that periodic temporal patterns do shed light on relationships between nodes and capture them, which is the clear distinction between DynamicTriad, HTNE and our model. We will show how both periodic and non-periodic temporal features contribute to the performance of our model in the experiments.

\subsection{Model Optimization}
In this section, we introduce our objective function that is able to synthesize both structural proximity and evolutionary patterns, followed by our optimization scheme. 

\subsubsection{Preserving Evolutionary Patterns}
In this section, we introduce the intuition underlying our temporal objective function, followed by its formal definition. 

Since the representation vectors preserve pairwise proximity between nodes, we propose that relative positions between node pairs reflect relationships between the nodes. For example, if $\|\mathbf{u}_1 -\mathbf{u}_2 \| = \|\mathbf{u}_3 - \mathbf{u_4}\|$ but $(\mathbf{u}_1 - \mathbf{u_2})\cdot (\mathbf{u}_3 - \mathbf{u}_4) = 0$, we would believe that node pairs $1, 2$ and $3, 4$ are identically adjacent in network space, yet edge $\langle 1, 2\rangle$ and $\langle 3, 4\rangle$ represent relationships that are scarcely correlated. 

Since we propose that evolutionary patterns can implicitly reveal relationships between node pairs, as illustrated by the example of colleagues and families, we would hence derive that, evolutionary patterns are also indicative of relative positions between pairwise nodes, and vise versa. 

The intuition introduced above can be implemented by an auto-encoding style objective function
\begin{equation}
L_{temporal} = \sum_{v_i \in V} \sum_{v_j \in N^t(v_i)} D\left( \mathbf{u}_i^t - \mathbf{u}_j^t, g\left(\mathbf{s}^t\left(v_i\right), \mathbf{s}^t\left(v_j\right)\right)\right),
\label{equ:temporal_loss}
\end{equation}
where $\mathbf{u}_i^t - \mathbf{u}_j^t$ implies the relative position between node pairs, $g(\cdot, \cdot)$ is some function aiming to interpret relative positions between node pairs through temporal features, and $D(\cdot, \cdot)$ measures a distance metric between two vectors. In particular, in our model, we set
\begin{equation}
    g\left(\mathbf{s}^t\left(v_i\right), \mathbf{s}^t\left(v_j\right)\right) = \sigma\left(\mathbf{W} \cdot \left(\mathbf{s}^t\left(v_i\right) \oplus \mathbf{s}^t\left(v_j\right)\right)\right),
\end{equation}
where $\oplus$ denotes concatenation, $\mathbf{W} \in \mathbb{R}^{d_1 \times 2 d_2}$ is a trainable matrix,  $\sigma(x)$ denotes the sigmoid function, and $D(\mathbf{x}, \mathbf{y}) = \cos(\mathbf{x}, \mathbf{y})$ denotes cosine similarity between vectors. 

\subsubsection{Preserving Node Proximity}
In this paper, we preserve proximity between pairwise nodes through an objective derived from DeepWalk, but it should also be noticed that EPNE is not restricted to specific models and can be generally incorporated to capture temporal information. We maximizes the likelihood that the context $v_j\in N^t(v_i)$ is observed conditional on the representation vectors $\mathbf{u}^t_i$ and $\mathbf{u}^t_j$. Our objective preserving proximity, accelerated by negative sampling, is defined as 

\begin{equation}
\begin{aligned}
\min L_{struct} &=  - \sum_{v_i \in V} \sum_{v_j \in N^t(v_i)} \bigg[\log \sigma\left(\left(\mathbf{u}_i^t\right)^T \mathbf{u}_j^t\right) \\
& + k \cdot \mathbb{E}_{v_n \sim P_n(v)} \log \sigma\left(-\left(\mathbf{u}_i^t\right)^T \mathbf{u}_n^t\right)\bigg].
\label{equ:structural_loss}
\end{aligned}
\end{equation}

\subsubsection{Overall Loss}
We notice that the structural objective possesses the property of rotational invariance, i.e. the objective does not change regardless of how the vector space is rotated, which 
poses potential threats to the learning of our model. As we model temporal features in different vector spaces between time steps, it is important that those spaces should be aligned such that we do obtain evolutionary patterns instead of random rotations of vector spaces between time steps. To enforce such restriction, we propose a loss function imposing smoothness in a weighted manner between adjacent time steps
\begin{equation}
L_{smooth} = \sum_{t=1}^T \sum_{i=1}^{|V|}\frac{\|\mathbf{u}_i^t-\mathbf{u}_i^{t-1}\|}{\|\mathbf{d}_i^t-\mathbf{d}_i^{t-1}\|},
\label{equ:smoothness_loss}
\end{equation}
where $\|\mathbf{d}_i^t-\mathbf{d}_i^{t-1}\|$ measures how the structure of node $v_i$ changes from time $t-1$ to time $t$ as defined in \cite{DBLP:journals/corr/abs-1802-06257}. 

We have the overall loss function of our model by summing these objectives
\begin{equation}
\min L = L_{struct} + \alpha L_{temporal} + \beta L_{smooth},
\label{equ:overall_loss}
\end{equation}
where $\alpha$ being the temporal weight and $\beta$ being the smoothness weight are hyperparameters to be tuned. 

We apply Stochastic Gradient Descent (SGD) to optimize the objective function. We learn representation vectors for nodes $v_i^t$ at each time step $t$ incrementally based on historical representation vectors, thereby restricting the number of parameters at each time step within $O(|V|)$. We summarize our learning algorithm in Algorithm \ref{alg:EPNE}.

\begin{algorithm}
	\caption{Incremental Learning Algorithm of EPNE}
	\label{alg:EPNE}
	\begin{algorithmic}[1]
		\Require Network snapshot at time $t$:  $G^t = \{V, E^t\}$
		\Require Embeddings in snapshot before time $t$: $\mathbf{u}_i^{t'}, t' < t$
		\Ensure Embeddings in snapshot at time $t$: $\mathbf{u}_i^t$
		\State Initialize embeddings randomly $\mathbf{u}_i^t$
		\For{each $v_i \in V$}
			\State $\mathcal{W}_i=\mathit{RandomWalk(G^t, v_i)}$
		\EndFor
		\For{$e=1$ to $num\_epoches$}
		\For{each sequence $\mathcal{W}_i$}
		    \State $v_i = \mathcal{W}_i[0]$, $v_j \in \mathcal{W}_i[1:w]$
		    \State Calculate frequency domain features $\mathbf{s}_{freq}^t(v_i)$ of node $v_i$ according to Equation \ref{equ:freq_features}
		    \State Calculate time domain features $\mathbf{s}_{time}^t(v_i)$ of node $v_i$ according to Equation \ref{equ:time_features}
		    \State Temporal features $\mathbf{s}^t(v_i) = \mathbf{s}^t_{time}(v_i) \oplus \mathbf{s}^t_{freq}(v_i)$ 
		    \State Calculate evolutionary pattern preserving loss $L_{temporal}$ according to Equation \ref{equ:temporal_loss}
		    \State Calculate structure preserving loss $L_{struct}$ according to Equation \ref{equ:structural_loss}
		    \State Calculate smoothness term $L_{smooth}$ according to Equation \ref{equ:smoothness_loss}
		    \State Overall loss $L = L_{struct} + \alpha L_{temporal} + \beta{L_{smooth}}$
		    \State Update embeddings $\mathbf{u}_i^t = \mathbf{u}_i^t - \frac{\partial L}{\partial \mathbf{u}_i^t}$ and $\mathbf{W}$
		\EndFor
		\EndFor
	\end{algorithmic}
\end{algorithm}

\section{Experiments}
In this section, we evaluate our model on several real-world networks on node and edge classification tasks. We first introduce our experimental setups, followed by quantitative and qualitative results.

\subsection{Experimental Setup}

\subsubsection{Datasets}
We employ the following datasets with temporal information, whose statistics are listed in Table \ref{tab:data_stats}. 
\begin{itemize}
\item \textit{High School} \cite{DBLP:journals/corr/MastrandreaFB15} is a social network collected in December, 2013. We consider students as nodes and active contacts collected by wearable sensors as edges. We split time steps by one-hour intervals. The labels of edges denote Facebook friendship. 
\item \textit{Mobile.} 
We build a social network from a mobile phone dataset collected in October 2010. We consider users as nodes and interactions as edges
. We split each time step by a two-hour interval. The labels of edges represent colleague relationships between users.
\item \textit{AMiner} \cite{zhou2018dynamic,ijcai2019-606} and \textit{DBLP} \cite{zuo2018embedding} are both co-author networks in which nodes represent researchers and edges represent co-authorship. For temporal settings, the datasets are split with four-year and one-year intervals respectively. The node labels represent research fields according to the conferences the author published his papers in. We take edge labels in AMiner as whether two researchers share the same field of interest.
\end{itemize}

\subsubsection{Baselines}
We compare our model with the following novel methods. Unless specified, these models are tested with their published codes as well as parameter settings mentioned in their original papers.  

\begin{itemize}
\item \textit{Static Skip-Gram models} including DeepWalk \cite{perozzi2014deepwalk}, node2vec \cite{grover2016node2vec} and LINE \cite{tang2015line}. For node2vec, we take $p = q = 0.25$.
\item \textit{Static Graph Neural Networks}. We take GraphSAGE \cite{hamilton2017inductive} as an example. We take unsupervised GraphSAGE without node features. We take 2-layer networks with a hidden layer sized 100. We adopt the neighborhood sampling technique with 20 neighbors to sample at each layer. 
\item \textit{HTNE} \cite{zuo2018embedding}.
It is a dynamic network embedding method modeling sequential neighborhood formation processes using Hawkes Process.
\item \textit{DynamicTriad} \cite{zhou2018dynamic}.
It is a dynamic network embedding method based on triadic closure processes and social homophily. We set $\beta_0 = 1$ and tested $\beta_1\in \{0.01, 0.1\}$ for optimal performances.
\end{itemize}
For static methods, we compress all temporal snapshots of the graphs into a ``stacked" static graph as mentioned in Definition \ref{defn:tempnet}, just like what  \cite{zuo2018embedding,zhou2018dynamic} did for static methods. 

For a fair comparison, the embedding dimensions are all set to $\mathbf{32}$. The number of paths for each node and the length of each path for DeepWalk, Node2Vec, and EPNE are all set to $10$. It is worth mentioning that, although Perozzi et al. \cite{perozzi2014deepwalk} took 80 paths per node, it was concluded redundant in our experiments as 10 paths per node have been capable of achieving similar performance. The window size for all random-walk based models is set to 5. In addition, we set $\alpha=1.0, \beta=0.01$ for all four datasets. We set the history length $h=45, 12, 12, 10$ and number of scales $L=3, 4, 4, 4$ for \textit{High School}, \textit{Mobile}, \textit{AMiner} and \textit{DBLP} respectively.

\begin{table}
\centering
\begin{tabular}{l|cccc}  
\toprule
Dataset  & $|V|$ & $|E|$ & \#temp. edges & \#$T$ \\
\midrule
High school & 327 & 5,818 & 188,508 & 45 \\
Mobile & 2,985 & 54,397 & 701,030 & 12 \\
AMiner & 11,056 & 70,217 & 308,886 & 32 \\
DBLP & 28,085 & 150,571 & 236,894 & 27 \\
\bottomrule
\end{tabular}
\caption{Dataset statistics.}
\label{tab:data_stats}
\end{table}

\subsection{Node Classification}
We first employ two networks, \textit{AMiner} and \textit{DBLP} for node classification experiments. The embedding vectors are trained using the whole graph, which are split into training and test sets for classification, using \textit{Logistic Regression} in \textit{sklearn} package. We vary the size of the training set from 10\% to 90\% of the whole dataset, such that the results would be indicative of embedding consistency. We repeat all experiments for 10 times and report their mean Macro-F1 scores. 

The results of node classification are shown in Table \ref{tab:node_classification_macro}. It can be shown that temporal embedding models generally outperform their static counterparts, which underscores that temporal features carry rich information that helps infer user communities. In addition, our model EPNE generally achieves the best performance, outperforming DeepWalk and beating its temporal counterparts, indicating that temporal information can be better captured and leveraged using evolutionary patterns learned by our model. In addition, it is also demonstrated that our model can generate embeddings that are consistent and discriminative enough with arbitrary amounts of training data.

\begin{table*}[t]
\centering
\begin{tabular}{c|ccccc|ccccc}  
\toprule
\hspace{20.9ex} & \multicolumn{5}{c|}{DBLP} & \multicolumn{5}{c}{AMiner} \\
\midrule
ratio of training & 10\% & 30\% & 50\% & 70\% & 90\% & 10\% & 30\% & 50\% & 70\% & 90\%\\
\midrule
EPNE	&	\textbf{0.6454}	&	\textbf{0.6454} 	&	\textbf{0.6472} 	&	\textbf{0.6505} 	&	\textbf{0.6553} 	&	\textbf{0.7799} 	&	\textbf{0.7972} 	&	0.7976 	&	\textbf{0.8055} 	&	\textbf{0.8189}    \\
\midrule
DeepWalk    &  0.5889 	&	0.6053 	&	0.6107 	&	0.6110 	&	0.6109 	&	0.7427 	&	0.7586 	&	0.7731 	&	0.7743 	&	0.7811   \\
LINE 	&   0.5437 	&	0.5628 	&	0.5658 	&	0.5660 	&	0.5701 	&	0.6618 	&	0.7172 	&	0.7283 	&	0.7325 	&	0.7417    \\
Node2Vec 	&	0.5769 	&	0.6014 	&	0.6089 	&	0.6102 	&	0.6106 	&	0.7599 	&	0.7819 	&	0.7851 	&	0.7842 	&	0.7838   \\
GraphSAGE & 0.5496  &   0.5859  &   0.5898  &   0.5893  &   0.5912     &   0.7035  & 0.7332  &   0.7365    & 0.7276  &   0.7430  \\
HTNE 	&	0.6037 	&	0.6188 	&	0.6225 	&	0.6248 	&	0.6245 	&	0.7757 	&	0.7938 	&	\textbf{0.8028} 	&	0.8007 	&	0.8071    \\
DynamicTriad & 0.6072 & 0.6099 & 0.6213 & 0.6275 & 0.6284   & 0.7681 & 0.7889 & 0.7893 & \textbf{0.8046} & 0.8048\\
\bottomrule
\end{tabular}
\caption{Macro-F1 of node classification on training sets of varying size on different datasets.}
\label{tab:node_classification_macro}
\end{table*}
\begin{table*}[!h]
\centering
\begin{tabular}{c|cc|cc|cc}  
\toprule
& \multicolumn{2}{c|}{High school} & \multicolumn{2}{c|}{AMiner} & \multicolumn{2}{c}{Mobile} \\
\midrule
&  Macro-F1 & Micro-F1  & Macro-F1 & Micro-F1  & Macro-F1 & Micro-F1 \\
\midrule
EPNE  	&\textbf{0.5467}	&\textbf{0.7039}  & \textbf{0.6197} & \textbf{0.7234}  &\textbf{0.8350}	&   0.8722 \\
\midrule
DeepWalk   &   0.5190    &	0.6829    & 0.4933 & 0.6857  	& 0.8257	&0.8629 \\
LINE 	&0.5157	&0.6898 	&0.4962	&0.6890 	&0.8323	&0.8711\\
Node2Vec    &0.4970	&0.6913 	&0.4417	&0.6819 	&0.8268	&0.8636\\
GraphSAGE   & 0.3951	&  0.6241   & 0.4311	&  0.6783	&0.7296	& 0.8035\\
HTNE 	&0.5138	& 0.7012 	&0.4213	&0.6766 	&0.8331	& \textbf{0.8734}\\
DynamicTriad 	&0.5102	&0.6933	&0.5098&	0.6856  	&0.8290	&0.8678\\
\bottomrule
\end{tabular}
\caption{Macro-F1 and Micro-F1 of edge classification on different datasets.}
\label{tab:edge_classification}
\end{table*}

\subsection{Edge Classification}
We employ three networks, High School, AMiner and Mobile for edge classification, testing whether temporal information does help infer relationships between nodes, and whether our objective facilitates inferring such relationships. It should be noticed that we use inconsistent datasets for node and edge classification because there are certain datasets where we have no access to node or edge labels. The representation vector of an edge is obtained by concatenating representation vectors of its two end nodes. Logistic Regression in sklearn package is alike used. We split each dataset into training and test sets with a ratio of 7:3. We also evaluate each model 10 times with different seeds and show the mean results, in terms of Macro-F1 and Micro-F1.

The results of edge classification are shown in Table \ref{tab:edge_classification}. As shown, our model is able to outperform all other baselines in High School and AMiner by $2\%$ and $7\%$, respectively. It is thus indicated that due to our explicit modeling of temporal patterns, they are preserved through representation vectors which helps identify node relationships, while other temporal methods, incapable of preserving such relationships, barely outperform their static counterparts in such tasks. 
As for Mobile, we assume that the performance of our model is compromised due to insignificant temporal characteristics, which can also be inferred by DynamicTriad and HTNE's inability to generate better representations. 

\begin{figure*}[t]
\centering
\begin{minipage}[b]{1.0\linewidth}
\subfigure[Temporal Feature] { 
\includegraphics[width=0.18\columnwidth]{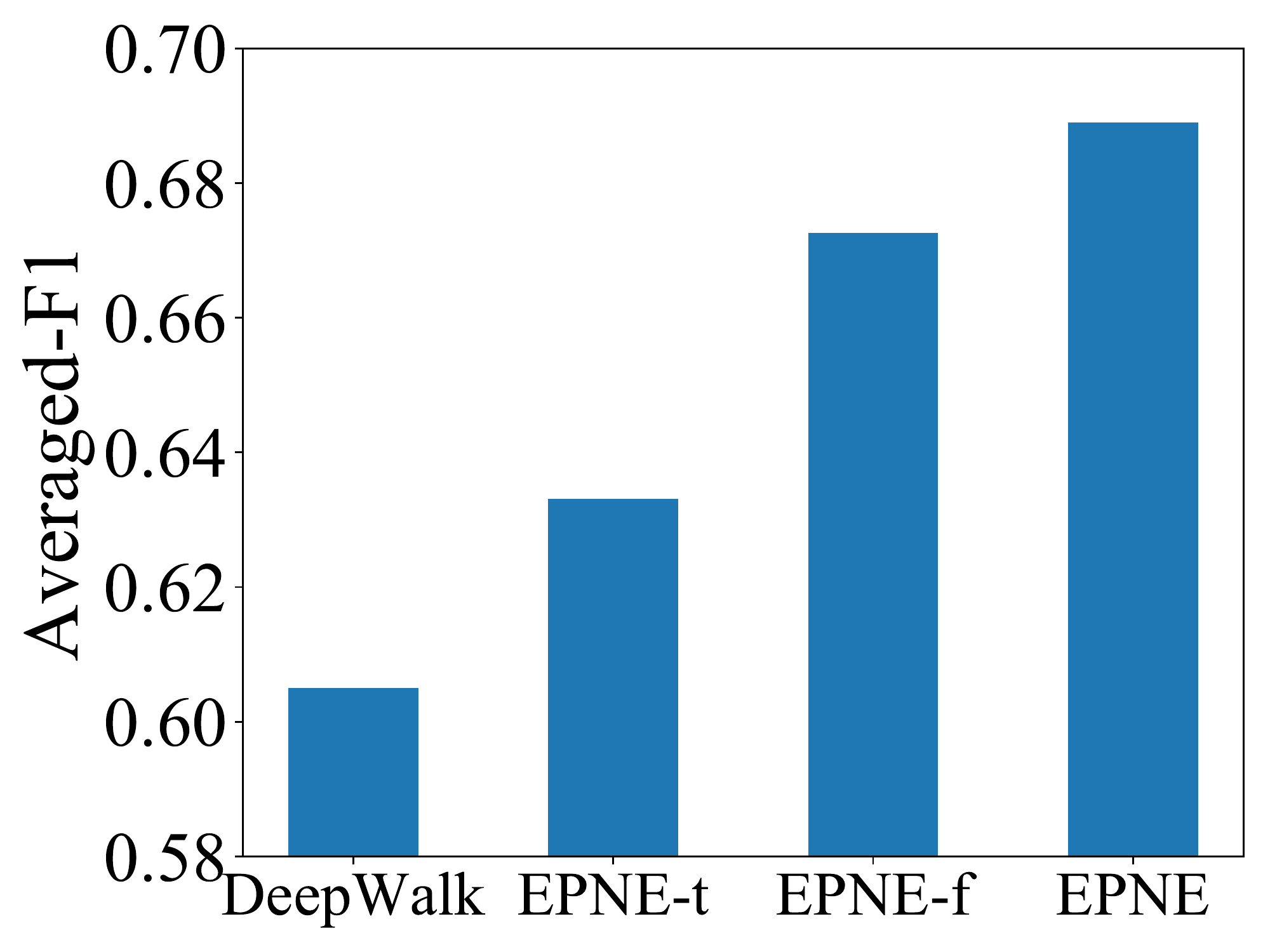} 
\label{fig:academic_feature}
}
\subfigure[Smoothness] { 
\includegraphics[width=0.18\columnwidth]{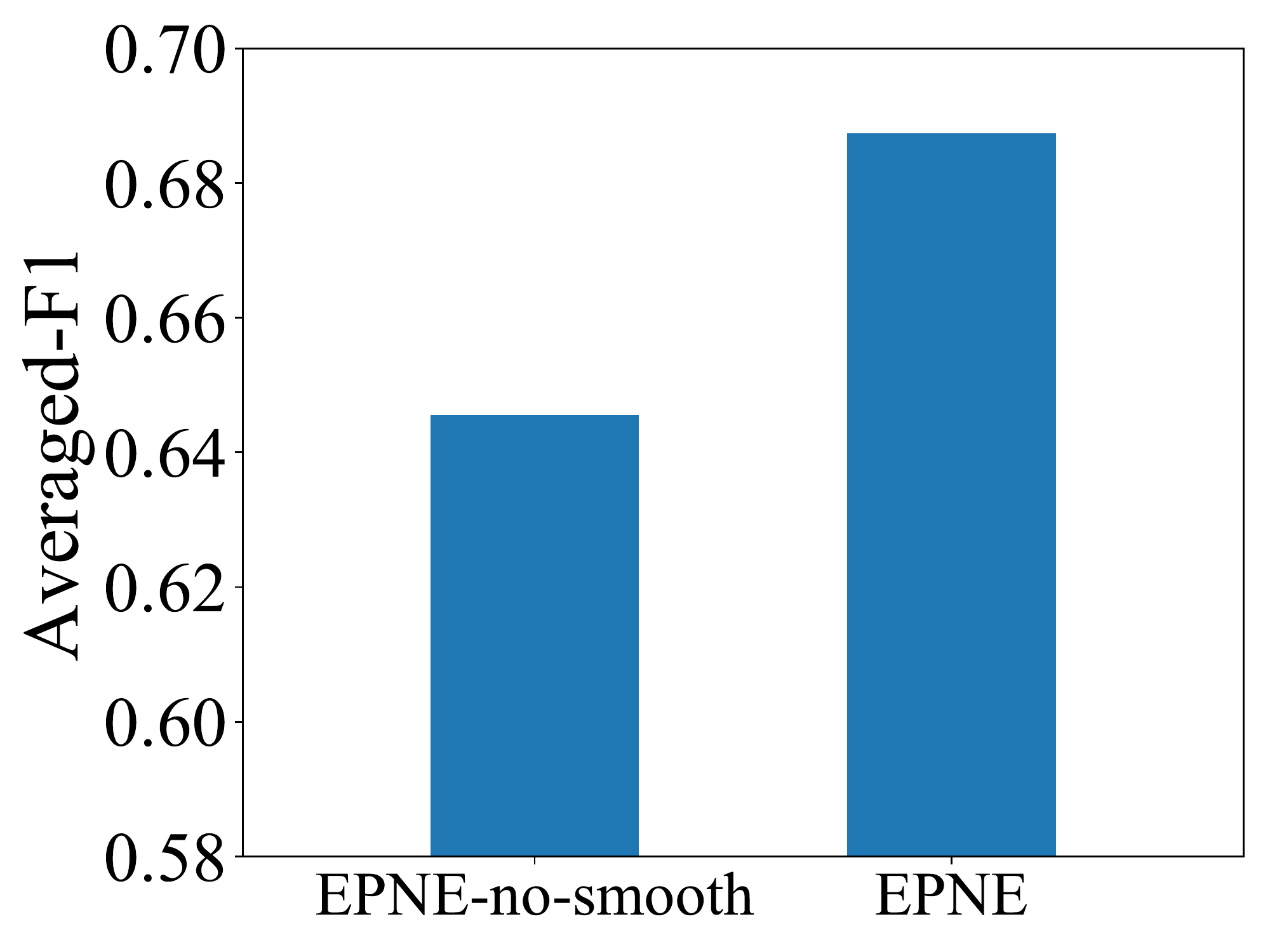} 
\label{fig:academic_smooth}
}
\subfigure[History Length] { 
\includegraphics[width=0.18\columnwidth]{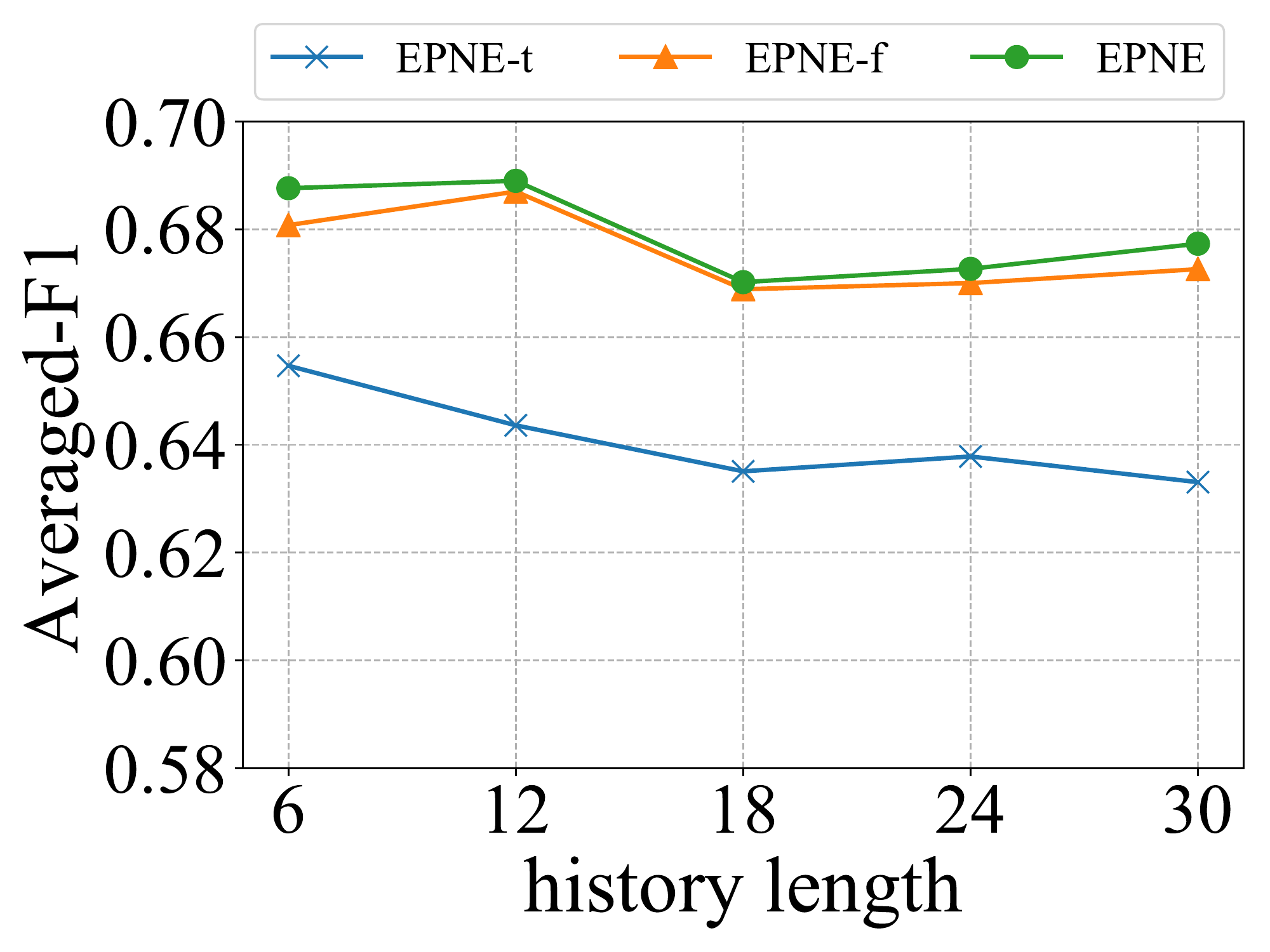} 
\label{fig:history_length}
}
\subfigure[Temporal Weight] { 
\includegraphics[width=0.18\columnwidth]{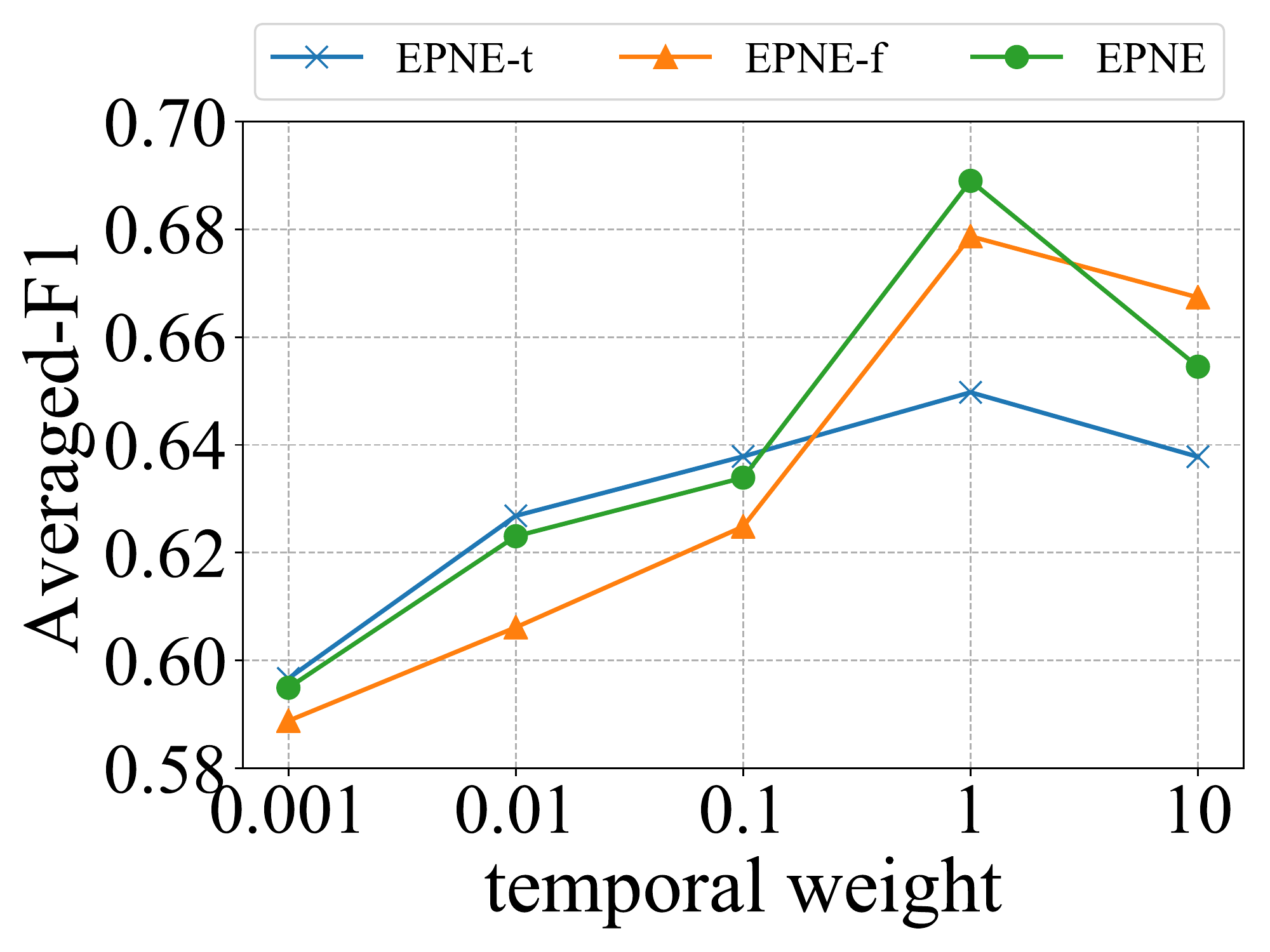} 
\label{fig:temporal_weight}
}
\subfigure[Scale Number] { 
\includegraphics[width=0.18\columnwidth]{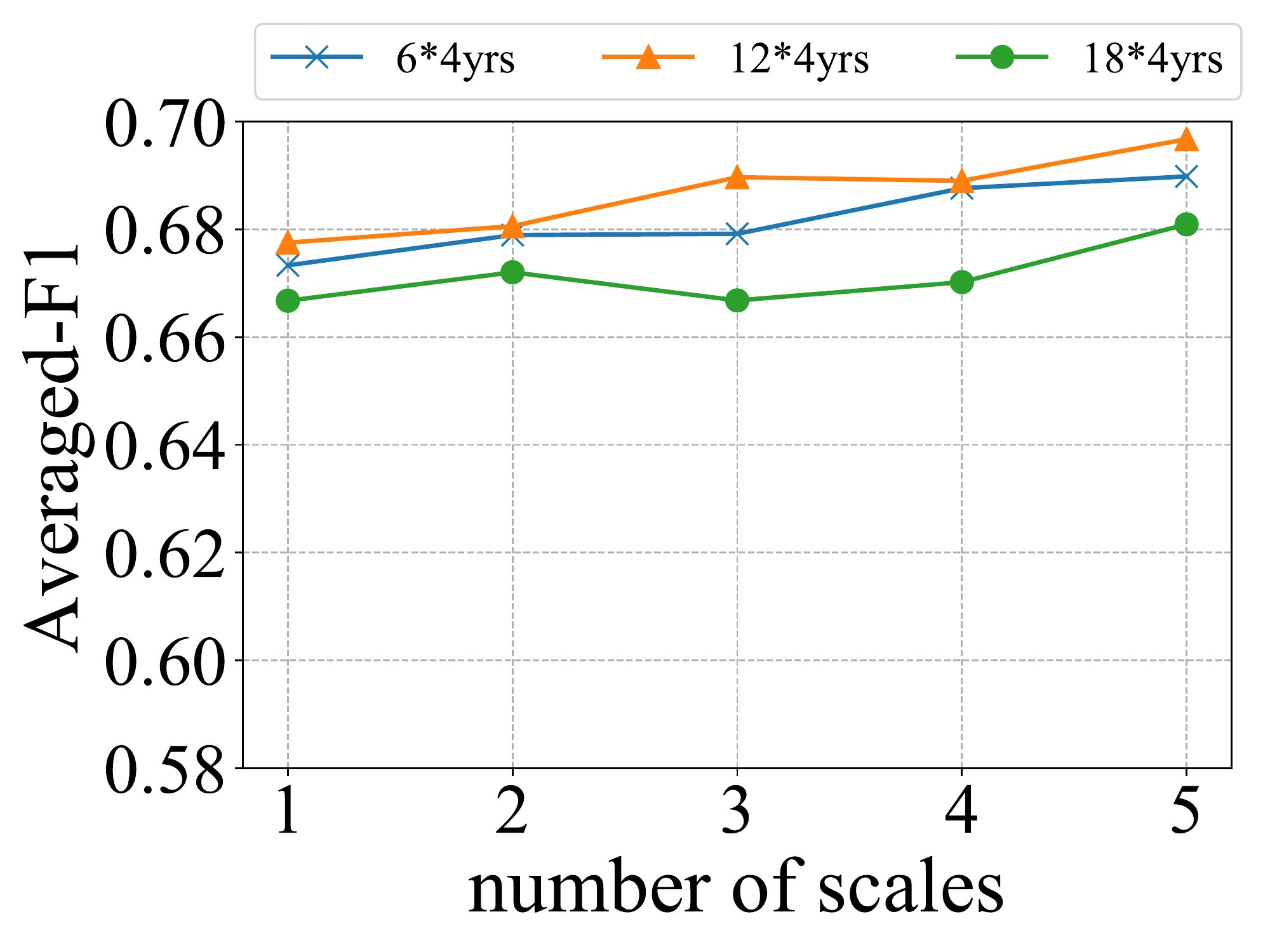} 
\label{fig:scale_num}
}
\caption{Parameter Analysis.}
\end{minipage}
\end{figure*}
\begin{figure*}[t]
\centering
\begin{minipage}[b]{1.0\linewidth}
\subfigure[DeepWalk] { 
\label{fig:vis_dw} 
\includegraphics[width=0.18\columnwidth]{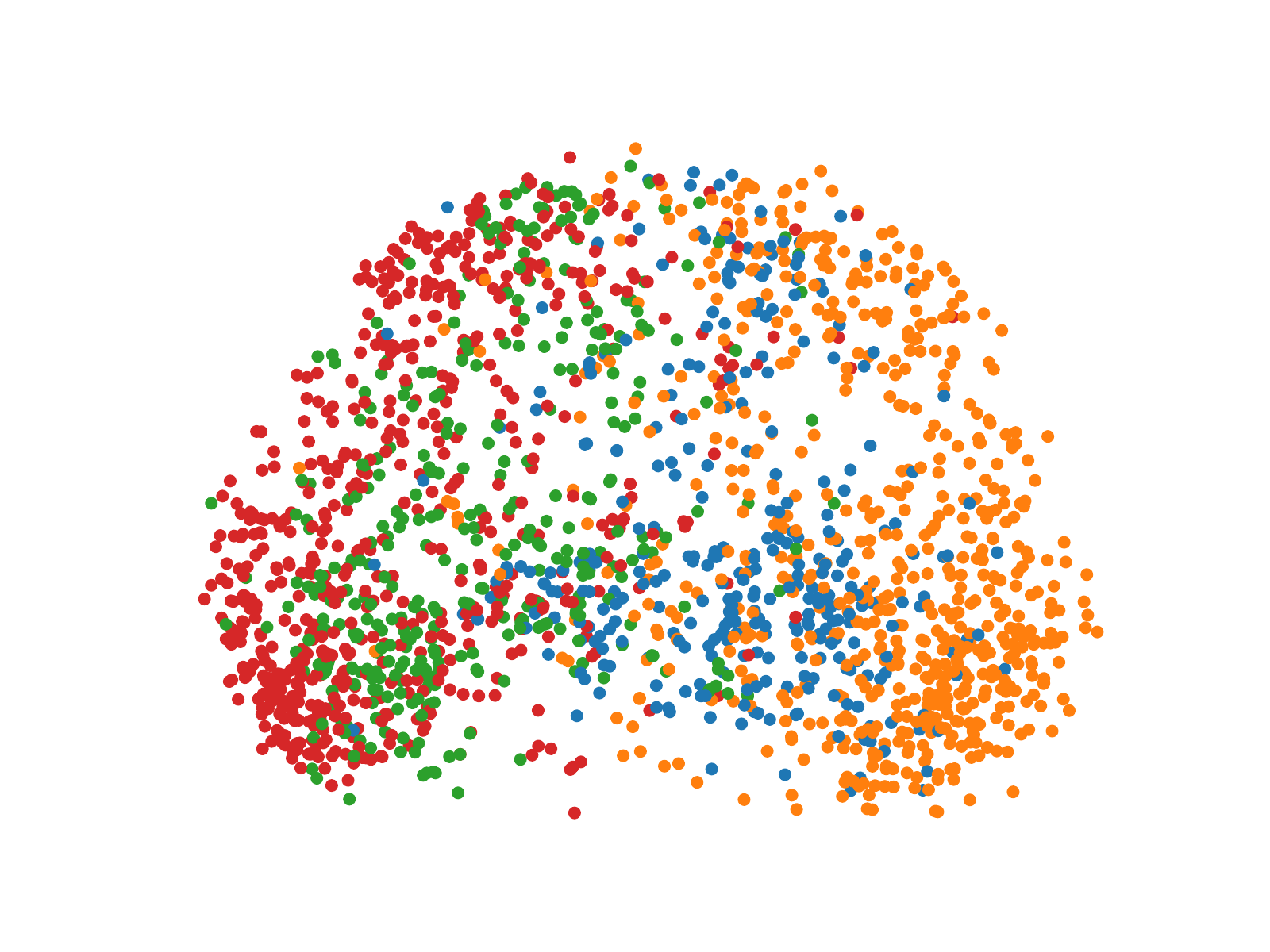}
} 
\subfigure[LINE] { 
\label{fig:vis_line} 
\includegraphics[width=0.18\columnwidth]{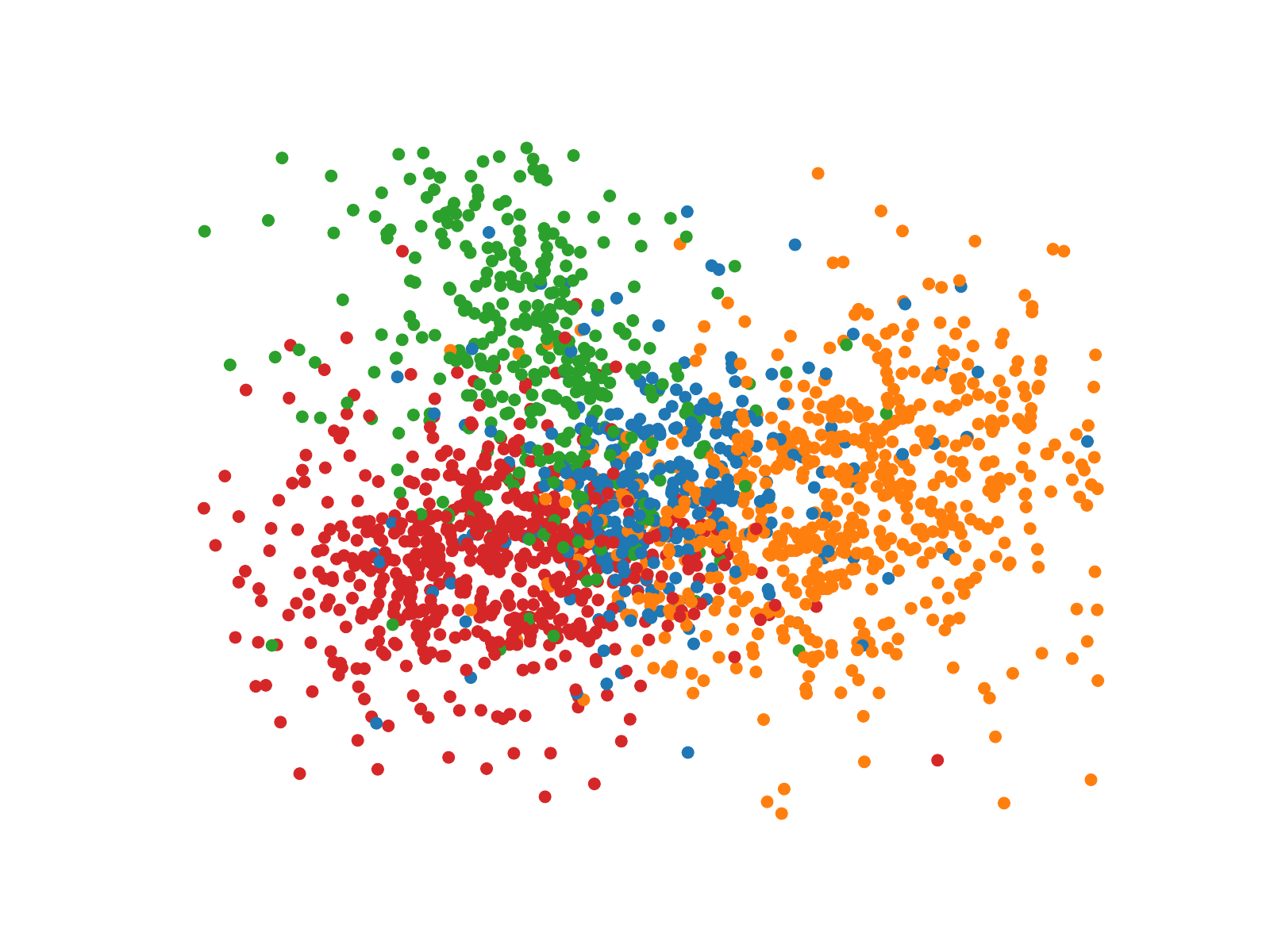}
} 
\subfigure[HTNE] { 
\label{fig:vis_htne} 
\includegraphics[width=0.18\columnwidth]{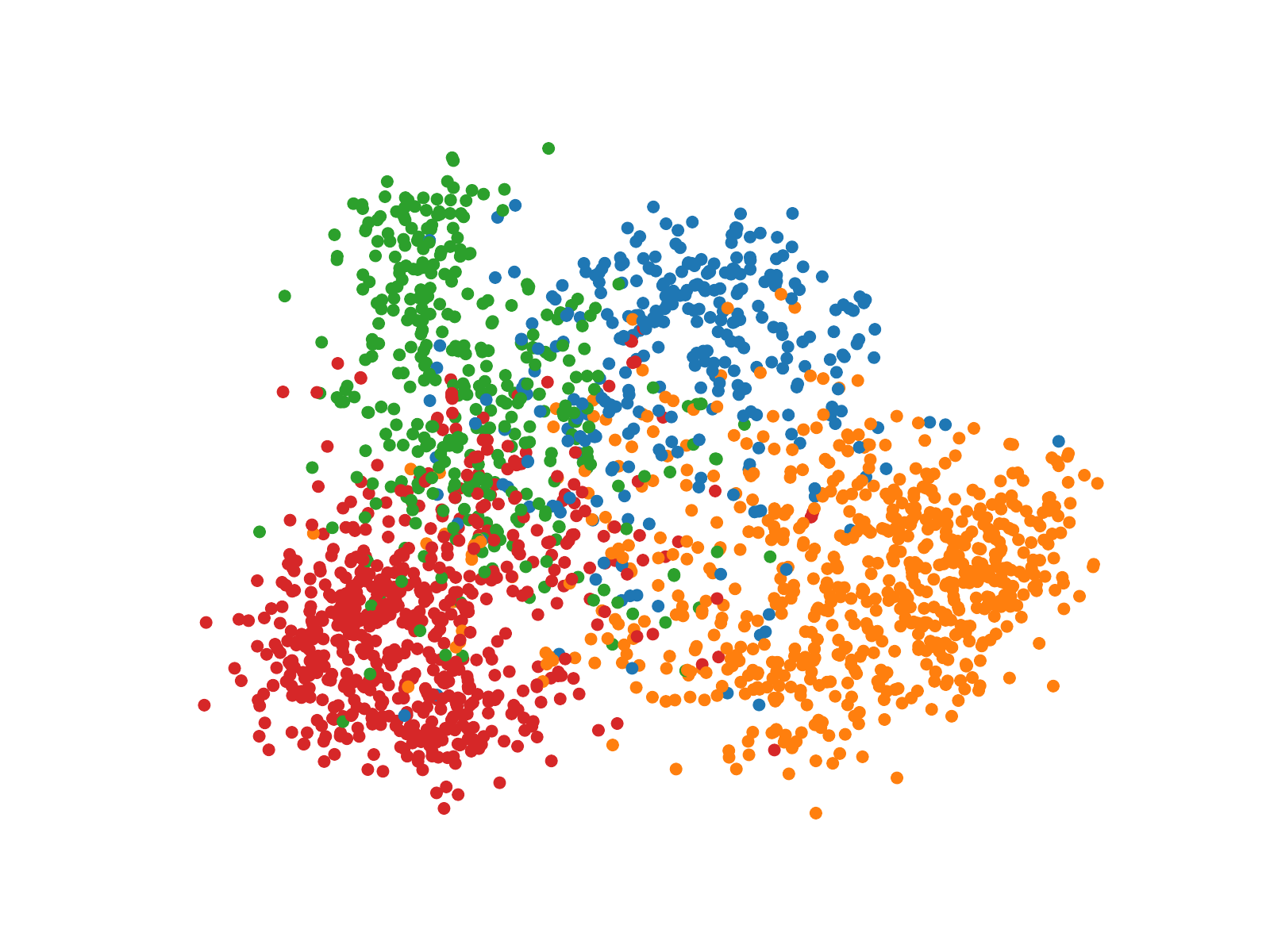}
} 
\subfigure[DynamicTriad] { 
\label{fig:vis_dyn} 
\includegraphics[width=0.18\columnwidth]{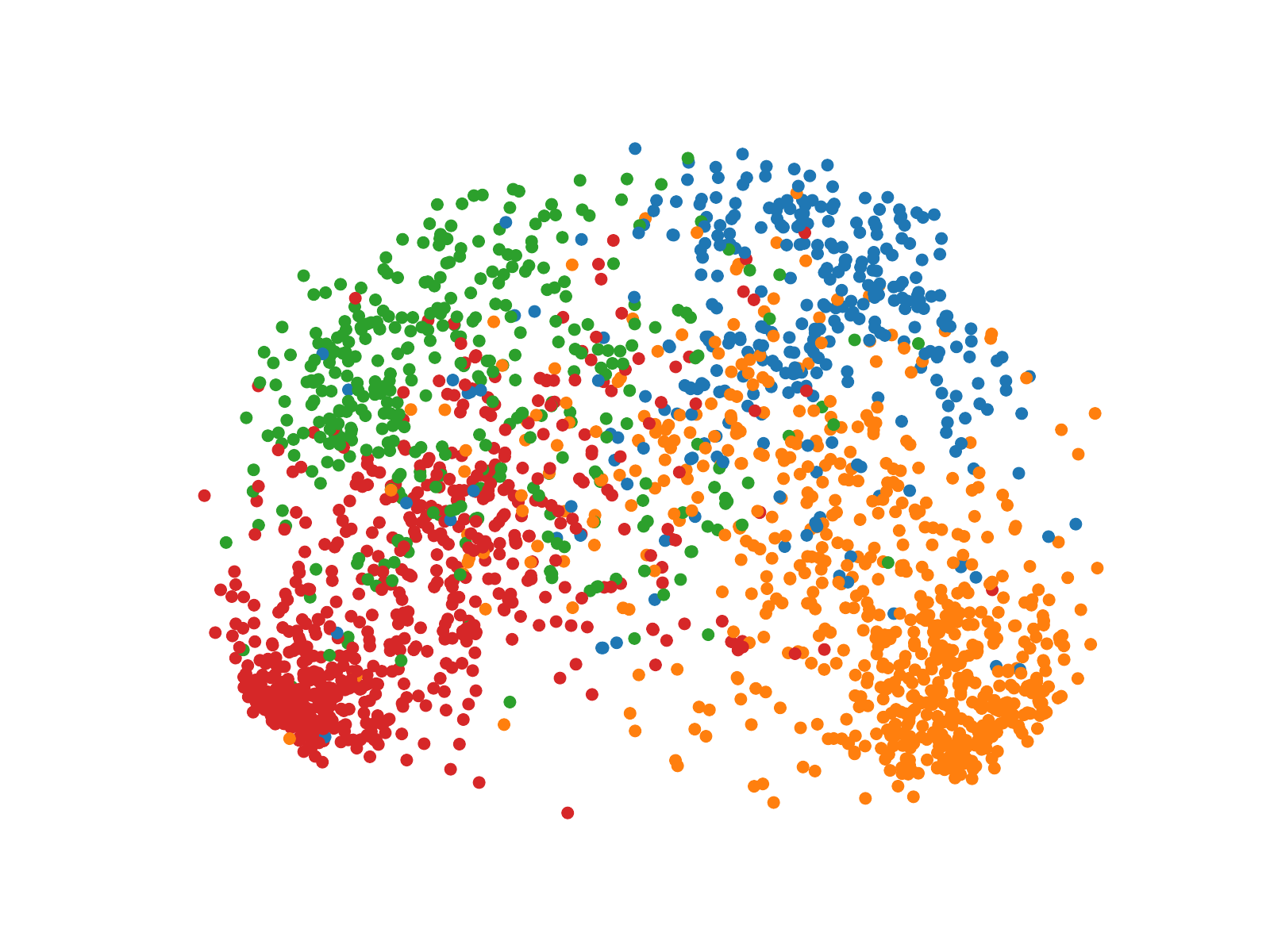}
} 
\subfigure[EPNE] { 
\label{fig:vis_tne} 
\includegraphics[width=0.18\columnwidth]{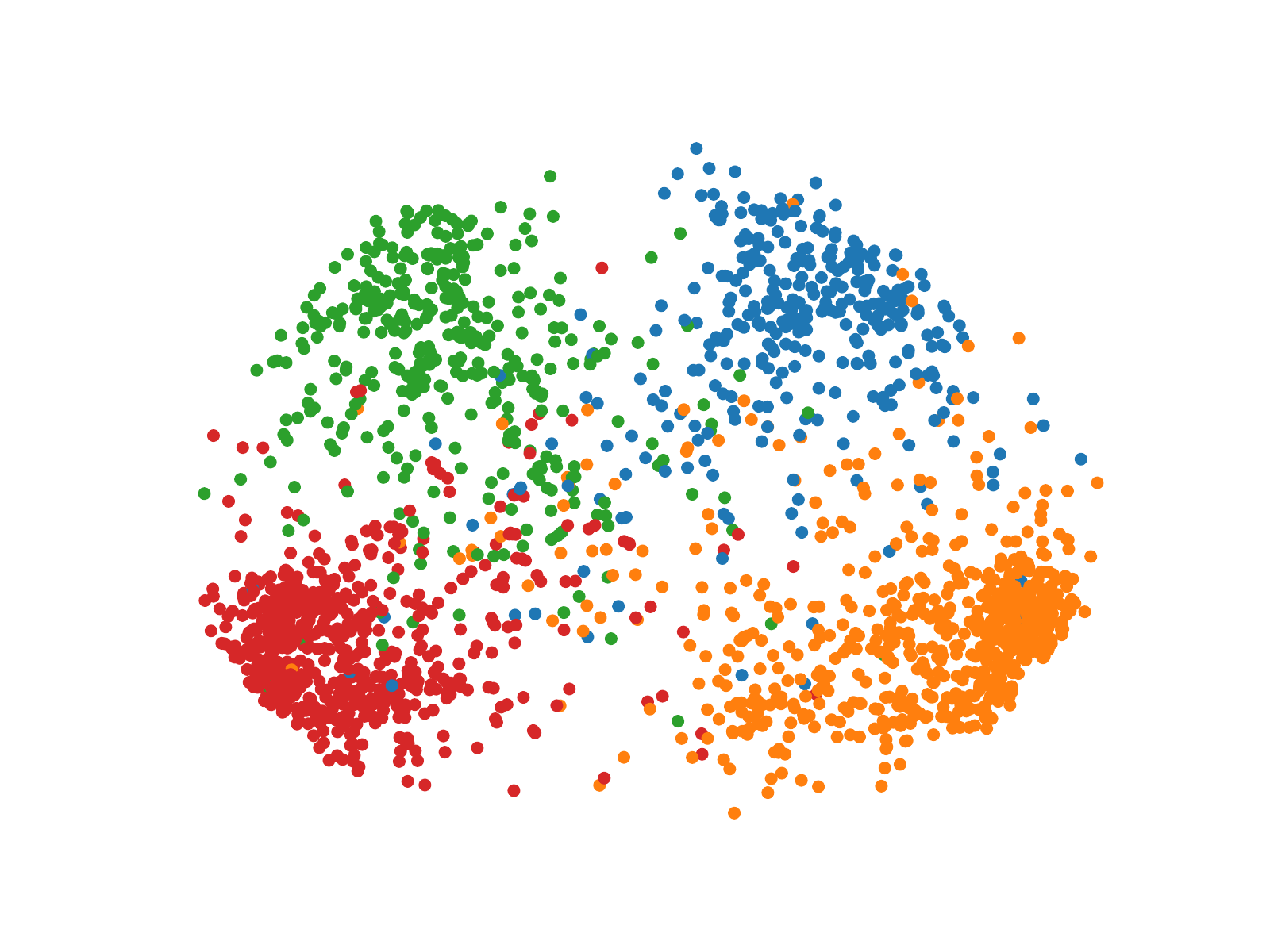}
} 
\caption{Visualization of embeddings generated by various algorithms in 2D space.} 
\label{fig:vis}
\end{minipage}
\end{figure*}

\subsection{Parameter Analysis}
In this section, we analyze our model's sensitivity to some parameters that are vital to the modeling of temporal information. We run our model on AMiner, in edge classification to show the performances. For brevity, we denote \textbf{EPNE-t} as our model with only non-periodic patterns, \textbf{EPNE-f} as our model with only periodic patterns, and \textbf{EPNE} as our complete model. All other parameters would be kept as default except those taken as parameters of the investigation. 

\begin{itemize}
    \item \textbf{Time-frequency Features:} 
    We carry out ablation studies on the temporal features we incorporate within our model. Specifically, we compare four models: one without any temporal features, or equivalently, DeepWalk, and three variants of our model: EPNE-t, EPNE-f, and EPNE. As shown in Figure \ref{fig:academic_feature}, our model combining both features outperforms models with either, demonstrating the utility of both features. Besides, EPNE-f performs better than EPNE-t, which indicates that periodic patterns are more indicative of relationships between nodes. 

    \item \textbf{Smoothness Term:} We analyze the performance of our model concerning the weight of smoothness term $\beta$. As shown in Figure \ref{fig:academic_smooth}, a distinctive gap between EPNE-no-smooth, the variant with $\beta = 0$ and EPNE is observed, demonstrating the effectiveness of the smoothness term in keeping embedding spaces aligned across all time steps.

    \item \textbf{History Length:} We study our model's performance with respect to the history length $h$. We set history length $h$ from 6 to 30 time steps. As shown in Figure \ref{fig:history_length}, the performance of our model climaxes at $h$ = 12 (48 years) before taking a huge plunge henceforth. It should not be surprising because co-authorship over 48 years ago sheds little light on the present. We thus conclude that a careful selection of $h$ is needed to ensure good performance.

    \item \textbf{Temporal Weight:}  We also study the performance with respect to the temporal weight $\alpha$. We test three versions of our model with $\alpha=$ 0.001, 0.01, 0.1, 1, 10. As shown in Figure \ref{fig:temporal_weight}, poor performance is observed when the temporal weight is either negligible or too large, leaving only one feature at work. 
    It is then concluded that our model performs the best with a moderate weight attached to the temporal objective, and as illustrated in the above dataset, a good empirical selection would be $a\approx 1$.

    \item \textbf{Number of Scales:} We analyze the influence of the number of scales $L$ on our performance with different history length $h$ in Figure \ref{fig:scale_num}. 
    We set $L$ ranging from 1 to 5, with $h$ set to 24, 48 and 72 years (or 6, 12 and 18 time steps) on \textit{AMiner}. It can be observed that, when more scales are considered, better performance is generally ensured. 
   
\end{itemize}

\subsection{Network Visualization}
We make qualitative evaluations of our models by visualizing embeddings learned from different models. In this section, we compare our model with a representative static model, DeepWalk, along with two temporal models, DynamicTriad and HTNE. We select 2000 researchers randomly from four research fields from the \textit{DBLP} dataset, whose embedding vectors are then projected into two-dimensional space using PCA and plotted using scatter plots. 

The plots for visualization are shown in Figure \ref{fig:vis} where dots colored blue, orange, green and red represent researchers from the four fields. It can be shown that the static network embedding method, DeepWalk (Fig. \ref{fig:vis_dw}) fails to distinguish nodes with different labels, mixing red dots with green ones. By modeling temporal information in networks, temporal models are able to map nodes from different fields with distinctive boundaries. In addition, as shown in Figure \ref{fig:vis_tne}, our model, EPNE, can project nodes with identical fields in a more condensed manner compared to DynamicTriad (Fig. \ref{fig:vis_dyn}) with respect to red and green dots, while minimizing overlapping areas across red, green and blue dots for compared to HTNE (Fig. \ref{fig:vis_htne}). 

\section{Conclusion}
In this paper, we analyze the temporal evolutionary patterns in real-world networks and demonstrated that such evolutionary patterns would contribute to obtaining more distinctive embedding vectors.  
We thus propose a novel network embedding model to learn representation vectors preserving not only static network structures, but also evolutionary patterns. As we observe that evolutionary patterns consist of periodic and non-periodic ones, different strategies are designed to learn time-frequency features, followed by a temporal objective to be jointly optimized along with structural objective. 
We conduct experiments on several datasets and the results in both node and edge classification affirmed our model's ability to generate satisfactory embeddings with both structural and temporal information.

\ack This work was supported by the National Natural Science Foundation of China (Grant No.61876006).

\bibliography{ecai}
\end{document}